\documentclass{article}

\usepackage{PRIMEarxiv}

\usepackage[utf8]{inputenc} % allow utf-8 input
\usepackage[T1]{fontenc}    % use 8-bit T1 fonts
\usepackage{hyperref}       % hyperlinks
\usepackage{url}            % simple URL typesetting
\usepackage{booktabs}       % professional-quality tables
\usepackage{amsfonts}       % blackboard math symbols
\usepackage{nicefrac}       % compact symbols for 1/2, etc.
\usepackage{microtype}      % microtypography
\usepackage{lipsum}
\usepackage{fancyhdr}       % header
\usepackage{graphicx}       % graphics
\graphicspath{{media/}}     % organize your images and other figures under media/ folder

\title{Auto Lead Extraction and Digitization of ECG Paper Records using cGAN
}

\author{ Rupali Patil, Bhairav Narkhede, Shubham Varma, Shreyans Suraliya and Ninad Mehendale}

\begin{document}
\maketitle

\begin{abstract}
 \item Purpose: An Electrocardiogram (ECG) is the simplest and fastest bio-medical test that is used to detect any heart-related disease. ECG signals are generally stored in paper form, which makes it difficult to store and analyze the data. While capturing ECG leads from paper ECG records, a lot of background information is also captured, which results in incorrect data interpretation.
    \item Methods: We propose a deep learning-based model for individually extracting all 12 leads from 12-lead ECG images captured using a camera. To simplify the analysis of the ECG and the calculation of complex parameters, we also propose a method to convert the paper ECG format into a storable digital format. The You Only Look Once, Version 3 (YOLOv3) algorithm has been used to extract the leads present in the image. These leads are then passed on to another deep learning model which separates the ECG signal and background from the single-lead image. After that, vertical scanning is performed on the ECG signal to convert it into a 1-Dimensional (1D) digital form. To perform the task of digitalization, we used the pix-2-pix deep learning model and binarized the ECG signals.
    \item Results: Our proposed method was able to achieve an accuracy of 97.4 \%.
    \item Conclusion: The information on the paper ECG fades away over time. Hence, the digitized ECG signals make it possible to store the records and access them anytime. This proves highly beneficial for heart patients who require frequent ECG reports. The stored data can also be useful for research purposes, as this data can be used to develop computer algorithms that are capable of analyzing the data.
\end{abstract}

% keywords can be removed

\section{Introduction}
\begin{figure*}[h!]
    \centering
    \includegraphics[width =\linewidth]{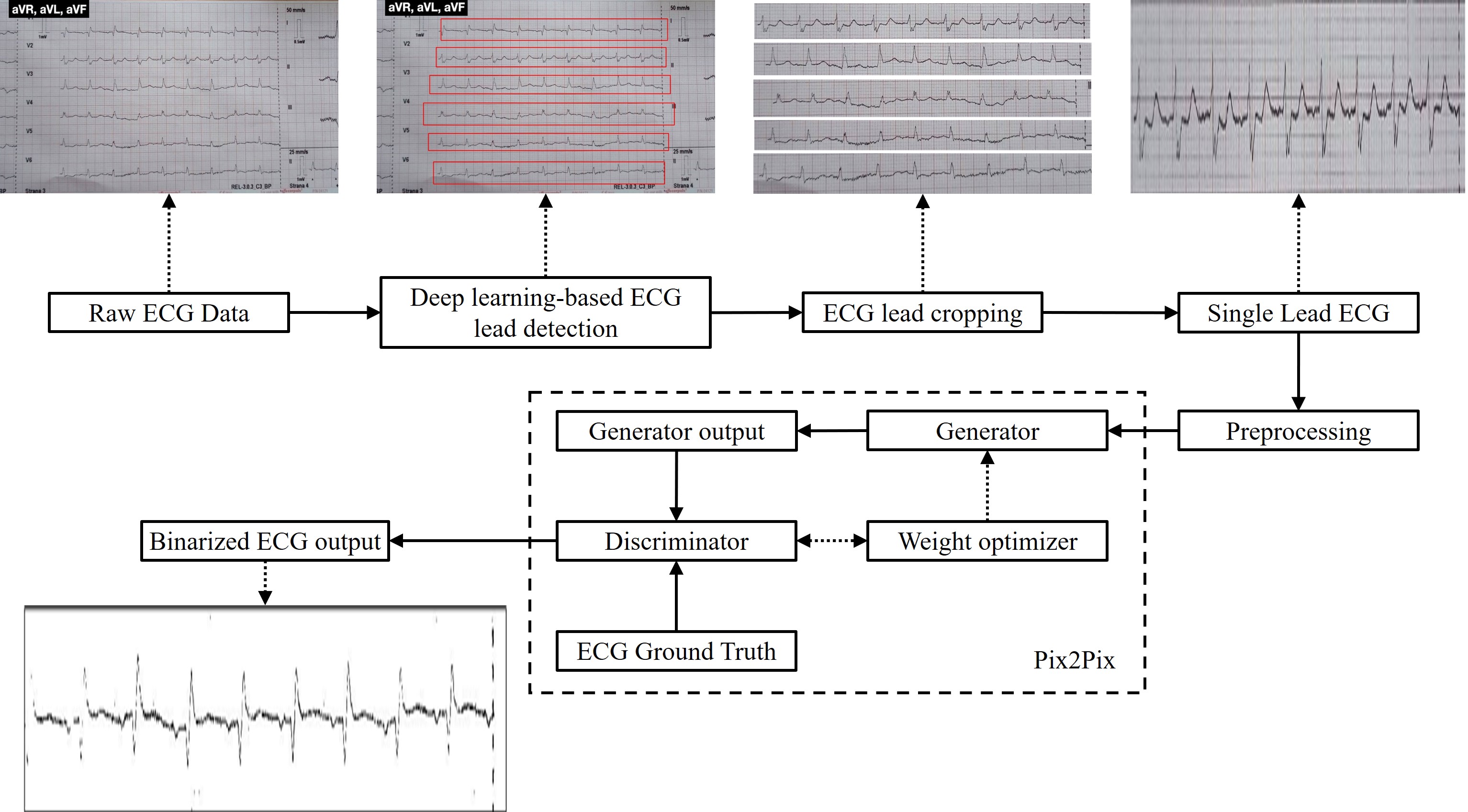}
    \caption{Flow diagram for deep learning-based digitization of ECG paper records. The data of each single lead ECG was segmented using deep learning methods. After segmentation, excess background was removed using a deep learning model. The digitized version of each single ECG image was generated using pix2pix. To get the 1D signal, the output images from the previous step were scanned in the vertical direction.}
    %\label{BcgImage}
    \label{fig1}
\end{figure*}

\begin{figure*}[h!]
    \centering
    \includegraphics[width = \textwidth]{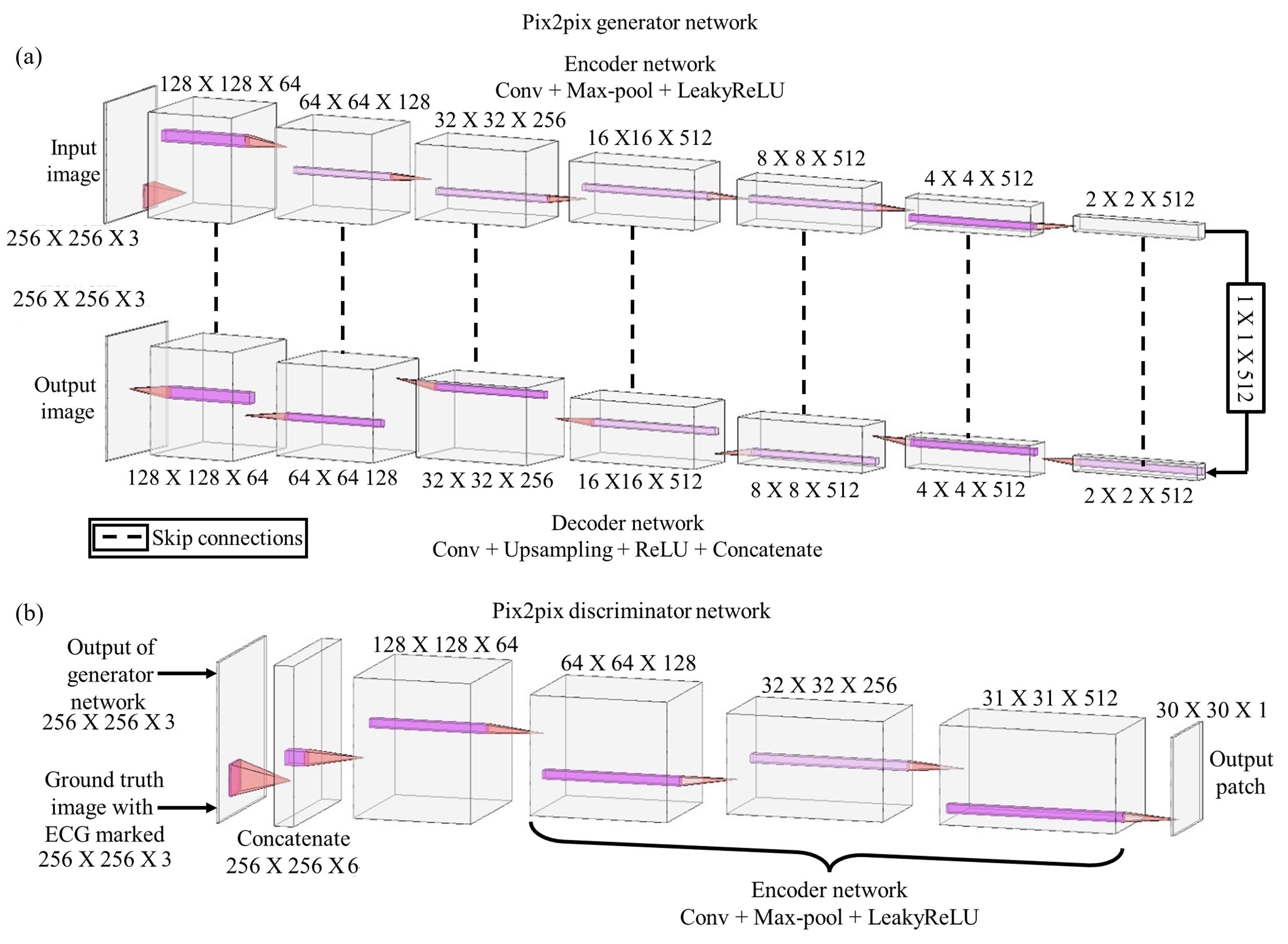}
    \caption{(a) The architecture of the pix2pix generator consists of an encoder and a decoder. An image of size 256 X 256 pixels with 3 RGB channels was passed to the encoder. It goes through a series of encoder blocks that extract information and compress the input image. The decoder consisted of a transposed convolution layer. The encoder encodes the image and translates the input image from its original dimensions of 256 X 256 X 3 pixels to 1 X 1 X 512 pixels, and then the decoder decodes the information back to the original dimension of the input image (256 X 256 X 3 pixels). (b) The discriminator takes two images as input: the output of the generator and the ground truth image. Each block of the discriminator contains a convolution layer, batch normalization layer, and leaky ReLU layer. Each 30 X 30 patch of the output classifies a 70 X 70 portion of the input image. It takes the input image as a concatenation of two images into one image of the shape of 256 X 256 X 6 pixels, and the output is of the size of 30 X 30 x 1 pixel. The network is said to be performing optimally when the discriminator classifies the ground truth image as the real image and the generator output as the fake image.}
    %\label{p2pgen}
    \label{fig2}
\end{figure*}

An Electrocardiogram (ECG) is the simplest and fastest biomedical test that is used to measure the activity of the heart \cite{price2010read}. Every heartbeat produces an electrical impulse signal. The ECG device makes use of electrodes placed on the patient's chest to capture heart activity. The electric signals generated by the heartbeats are captured by the electrodes. This method is useful to diagnose heart-related conditions \cite{hagiwara2018computer}. The ECG picture is a 1D graph with voltage on the Y-axis and time on the X-axis. The ECG thus gives different information, such as the frequency and strength of heartbeats. This allows doctors to diagnose the patient's heart health.

In the modern world, digital equipment is preferred over analogue ones as this equipment produces high-quality output and also provides flexibility in storage and future analysis. Digital technologies for signal acquisition, analysis, and diagnosis in the healthcare industry have emerged as an edge over the past. ECG recorders which capture the signals digitally are available, but the majority of the manufacturing companies produce machines that give a print of the ECG signal on paper \cite{prim2021data,tabassum2020numerical}. These paper copies, thus, cannot be stored as digital copies. One of the major drawbacks of paper ECG is that, after a certain time, the tracings of the ECG plots fade away. This problem makes it difficult for doctors or medical experts to acquire knowledge of a patient’s medical history. Digitized ECG can solve this problem, but despite having devices that can record and store ECG data digitally, most clinics and labs prefer paper ECG machines due to their low cost \cite{irianto2019low,prim2021data}. The advantage of digitised ECG is that they are easy to store, transmit, and retrieve. They can also be quickly interpreted and complex ECG parameters can be calculated by employing computer algorithms \cite{kligfield2007recommendations}. Digitally storing ECG data will also allow the use of various Artificial Neural Network (ANN) algorithms to analyze the signals and perform classification on them \cite{sao2015ecg}.

Hence, we propose a tool that can convert the ECG data on paper into a digital format with accurate waveform extractions using deep learning. We propose a model that uses the You Only Look Once (YOLO) deep learning algorithm to extract all 12 individual leads from a 12-lead ECG image captured by a camera. Each lead of an ECG signal from a 12-lead ECG format conveys unique information. The duration of 12-lead ECG signals considered was 5 secs. To obtain a standard 12-lead ECG, electrodes are placed on the limbs and the chest. The purpose of using multiple electrodes is to monitor the body's electrical signals from different perspectives. Our approach is also useful to remove the unwanted background information captured by the camera. The acquired single lead ECG image with its ground truth is then passed to the pix2pix model \cite{isola2017image} for binarization of ECG. The image with only an ECG signal is then vertically scanned to extract a 1D ECG signal, which can be used for further analysis. 

\section{Methodology}

\begin{figure*}[h!]
    \centering
    \includegraphics[width = \textwidth]{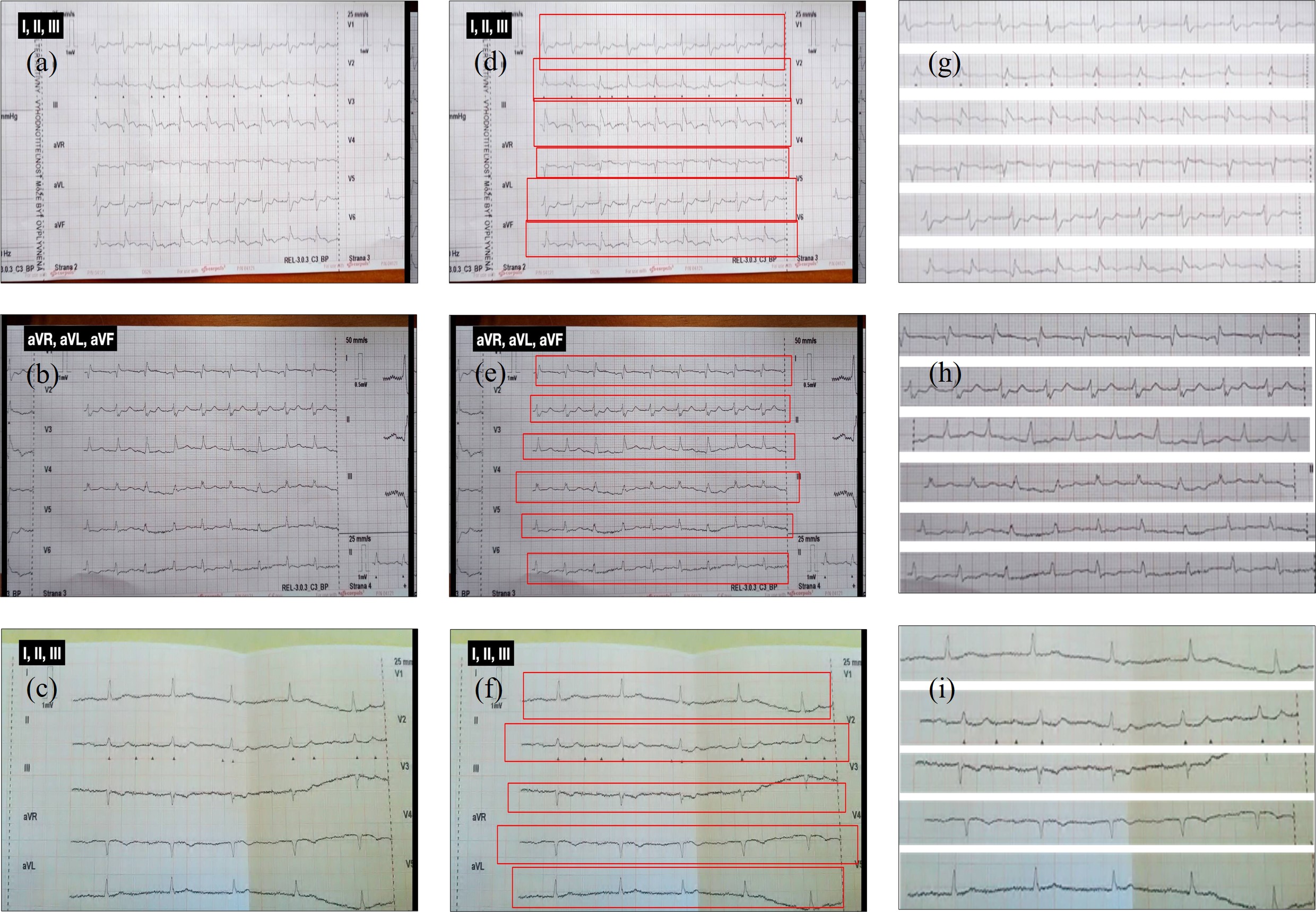}
    \caption{(a to c) Input images that were fed to the deep learning-based YOLOv3 model. (d to f) The bounding boxes drawn on each lead detected by YOLOv3. (g to i) The cropped lead images.}
    %\label{ROI}
    \label{fig3}
\end{figure*}

\begin{figure*}[h!]
    \centering
    \includegraphics[width = \linewidth]{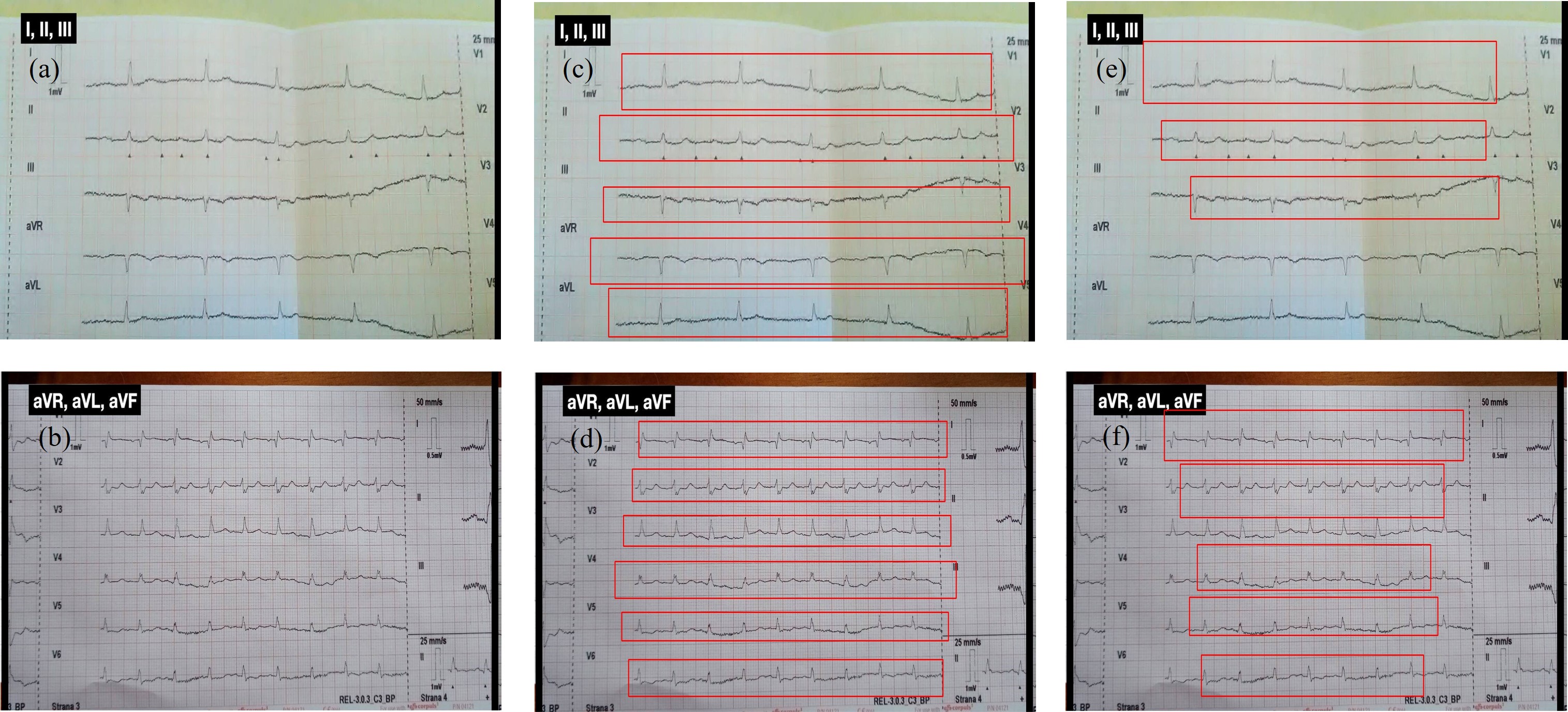}
    \caption{(a and b) Input images that were fed into the YOLOv3 and YOLOv4 models. (c and d) The bounding boxes drawn on each lead detected by YOLOv3.  (e and f) The bounding boxes drawn on each lead detected by YOLOv4. It can be observed that the YOLOv3 algorithm (c and d) detects the ECG signals more accurately than the YOLOv4 (e and f).}
    %\label{Compv3v4}
    \label{fig4}
\end{figure*}

Figure \ref{fig1} depicts a diagrammatic depiction of our proposed deep learning-based approach for digitising ECG from paper records. Initially, the proposed system auto-cropped the leads from the ECG picture before extracting the ECG signal. Deep learning was used to segment each lead's image, and then OpenCV was used to crop each lead. The cropped images were fed as input to a deep learning-based algorithm to binarize and eliminate the background from the ECG image. To get the 1D signal out of the output pictures from the previous phase, vertical scanning was used. 

\subsection{Image Acquisition}
Paper copies of the ECG results were provided by Saidhan Hospital and STEMI Global. The duration of 12-lead ECG signals considered was 5 secs. The first step was to capture the paper records with a camera and turn them into images. The file size of each image was between 1 megabyte and 2 megabytes. The resolution of the acquired images was 2880 x 2160 pixels. The database of the proposed system comprises ECG images acquired by the camera of the Samsung Galaxy S21+, as well as 12 lead ECG data records in the form of PDF files (Model: ECG600G). The ECG images were digitised using an automated digitizer machine (3040D) and Graph digitizer version 2.26 at Ninad's Research Lab, Thane, India. The values obtained from the automated digitizer machine (3040D) and the graph digitizer version 2.26 were converted to RGBA-formatted ECG signals.

\subsection{Region of Interest Extraction}
The ECG records were in the form of images. The first step of this methodology was to extract each lead from the images. Deep learning was used for this process. The proposed system used YOLOv3 \cite{redmon2018yolov3}, which was an object detection algorithm. The database created had a total of 53 images, of which were divided into two categories, training, and testing dataset. The training dataset contained 48 images, and the testing dataset contained five images. Since YOLOv3 was a supervised model, each lead in the training images had to be labeled. The leads were labelled in the form of a bounding box. LabelImg \cite{labelimg} was used to create a bounding box on every lead in the dataset. LabelImg is a graphical image annotation application developed in Python with a graphical interface built using Qt. LabelImg supports the YOLO format in which annotations are stored in a text file, and it also supports the Pascal VOC format in which annotations are stored in XML. In YOLO format, the annotations are stored in text format with the same filename as that of the image. For each label of a lead, i.e., bounding box, five parameters were generated, out of which the first parameter was the class number (in our case, the class was ‘lead’). The next two parameters were the bounding box's X and Y coordinates. The final two parameters specified the bounding box's width and height.  

Once all the images in the training dataset were labeled, the training of the proposed system was started. Due to the small size of the dataset (53 images), it was necessary to train the proposed system with a pre-trained model, which was darknet53.conv.74. Darknet53.conv.74 consisted of the initial weights for YOLOv3 that were trained on the ImageNet dataset. Darknet53.conv.74 had 75 CNN layers (convolutional layers) and 31 other layers (shortcut, route, upsample, YOLO), which added up to 106 layers in total. 

The training was performed with a batch size of 32 and 8 subdivisions, on Google Colab with a Tesla T4 Graphics Processing Unit (GPU) having 16GB of Video Random Access Memory (VRAM). The training was performed for 5056 iterations. Around 5056 iteration, our Mean Average Precision and F1 score started to stabilize. After 5056 iterations, the best weights generated were considered by comparing the Mean Average Precision (mAP@50\%) metric.

The final trained model had given output in the form of data about the bounding box, which was a single-column matrix. The matrix contains the parameters of the predicted bounding box, which are, Pc-confidence of the prediction, Bx and By-coordinates of the centre of the bounding box, Bw and Bh-height and width of the bounding box, and C1-class of the predicted bounding box. All these parameters were extracted using the Deep Neural Network (DNN) in the OpenCV library. The leads were then cropped using the bounding box coordinates obtained using the DNN library in OpenCV. The cropped leads were then passed on to the next stage, which is binarization.  

\subsection{Binarization using Pix2Pix}

Pix2Pix is a Conditional Generative Adversarial Network (cGAN) for image-to-image translation. GAN was a generative model that had two deep neural networks, a generator, and a discriminator. The generator was a modified U-Net, which consisted of two components, an encoder and a decoder. Each encoder consisted of convolutional layers, batch normalisation layers, and the Leaky ReLU activation function. Each decoder consisted of transposed convolutional layers, batch normalisation layers, dropout layers, and the ReLU activation layer. 
Every component in the discriminator network consisted of convolutional layers, batch normalisation layers, and the Leaky ReLU activation function. The discriminator received two input images: the target image (ground truth image) and the output of the generator. 
The generator was used to generate new images, which were then classified by the discriminator as real or fake. Then the weights of the generator and discriminator were adjusted during training. Training and weight adjustment of two deep neural networks were carried out separately. The performance of the generative model and the discriminative model improves with each epoch. The first step consisted of data preparation for the pix2pix model. There are many types of GANs, like Deep Convolutional GANs (DCGANs), Conditional GANs (cGANs), StackGANs, InfoGANs, Wasserstein GANs(WGAN), etc. Among all these types of GANs, cGANs could generate more detailed images if certain additional information was presented to them along with the raw input data. Hence, the cGAN model was provided with the image of a recorded ECG signal along with its corresponding digitised values. These corresponding digitised values were referred to as the ground truth. The digitised values for each ECG image in the dataset were manually traced and recorded in the RGBA format. The cGAN model used the ground truth as labels along with the raw input data to gain better insights. Hence, cGANs were suitable for binarization in the proposed method.  

Figure \ref{fig2} (a) shows the architecture of the Pix2Pix generator consists of an encoder and a decoder. The encoder was fed a 256 x 256 pixel image with three RGB channels. It passes through a series of encoder blocks, extracting information and compressing the input image. The decoder was made up of transposed convolution layers. The encoder encodes the image and converts it from its original dimensions of 256 X 256 X 3 pixels to 1 X 1 X 512 pixels, which the decoder then converts back to the original dimensions of the input image (256 X 256 X 3 pixels). The layers in the generator model embed information into an image. Since it can easily overfit, U-Net also introduced skip connections from the encoder to the decoder. The skip connections were concatenated in the encoder before going to the decoder. Encoder blocks contain a convolutional layer, a BN norm layer, and LeakyReLU activation. The decoder was composed of a convolution transposed. The decoder had convolutional layers, upsampling layers, a ReLU activation function, batch normalisation layers, and concatenation layers. 

Figure \ref{fig2} (b) shows the architecture of the discriminator network of pix2pix. The discriminator received two images as input: the generator output and the ground truth binarized ECG image. The convolution layer, batch normalisation layer, and leaky ReLU layer were all present in each discriminator block. Each 30 X 30 output patch identified a 70 X 70 part of the input image. It accepted an input image that is a concatenation of the two input images into one image with the dimensions of 256 X 256 X 6 pixels and gave an output image with the dimensions of 30 X 30 X 1 pixel. When the discriminator classifies the ground truth picture as the real image and the generator output as the false image, the network is said to be functioning optimally.

The digitised versions of 101 images were present in the dataset we used in our pipeline. Out of the 101 image pairs, 76 pairs were used as training datasets and 25 pairs were used as testing datasets. Before training of the pix2pix model, ECG images were processed by sharpening them using the matrix and filter2D operations of OpenCV. The contrast of the ECG image was then balanced using a Gaussian blur technique. The processed ECG image and its ground truth were used for training the pix2pix model. This model was designed using TensorFlow. The model was trained on 50000 steps with a batch size of 1, on Google Colab with a Tesla T4 GPU having 16GB of VRAM. For testing of the model, ECG images were first sharpened using the matrix and filter2D operations of OpenCV, and then Gaussian blur was used to balance the contrast of the ECG image. ECG images were then tested using the trained model. For the extraction of 1D ECG signals, the generated images were further processed, and then filtering of pixels was performed to extract the signals. The binarized, processed single lead ECG images were then vertically scanned \cite{patil2015digitization} to extract the 1D signals.

\begin{figure*}[ht!]
    \centering
    \includegraphics[width = \linewidth]{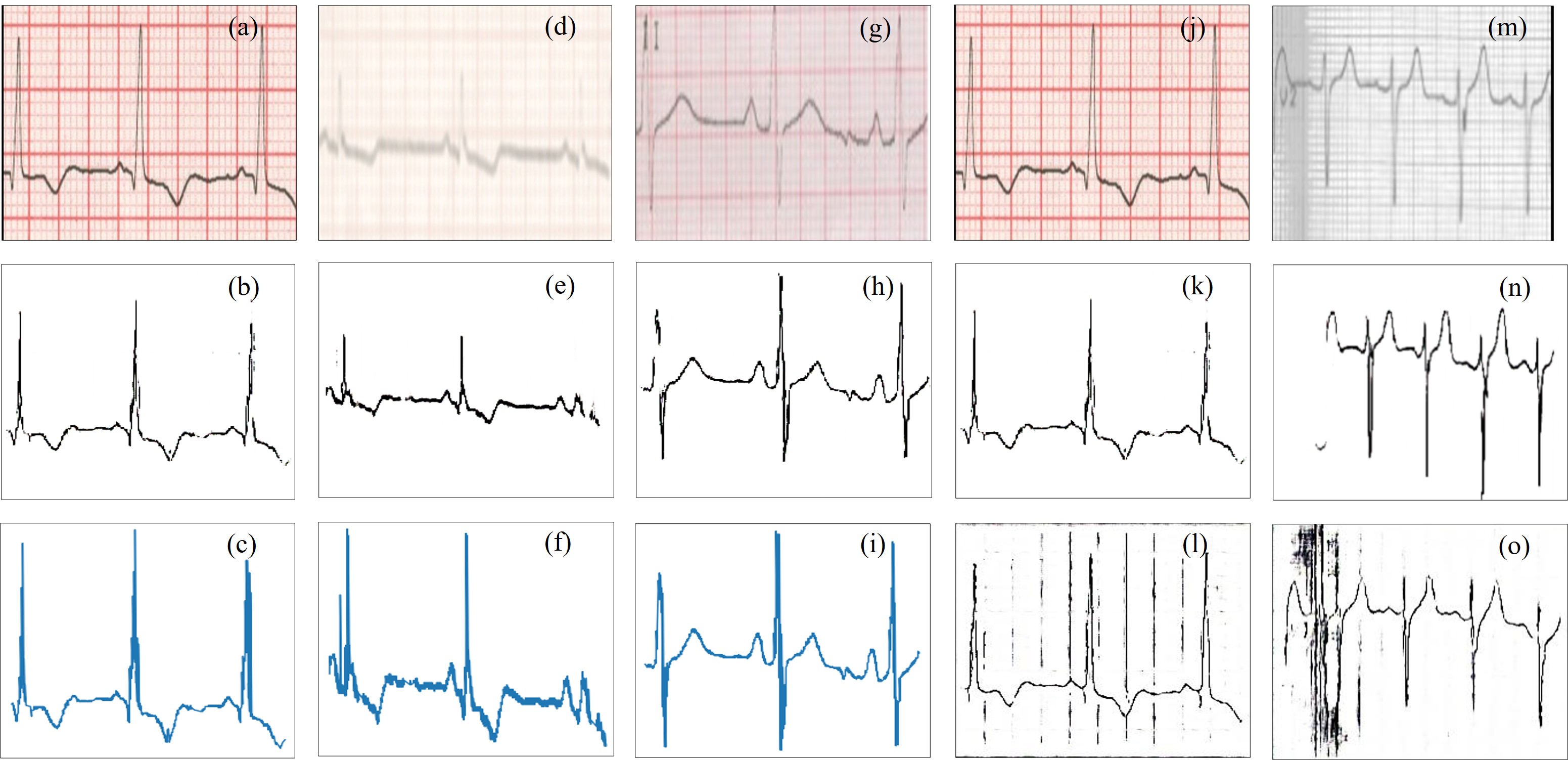}
    \caption{(a) Single lead ECG image obtained after detecting the region of interest (b) ECG image was preprocessed and binarized using the pix2pix model. (c) Vertical scanning is used to extract a 1D signal from a binarized ECG. (e-f) Extraction of 1D signal from the blurred ECG image shown in (d). (h-i) Extraction of 1D signal from an ECG image where contrast is very low and some signal is almost overlapping with the background shown in (g). The binarized image generated by pix2pix as shown in (k) and the binarized image generated by CycleGAN as shown in (l) of the clean ECG image shown in (j). (n) Pix2pix-generated binarized image and (o) CycleGAN-generated binarized image of the grayscale ECG image shown in (m). }
    %\label{p2pcgancomp}
    \label{fig5}
\end{figure*}

\subsection{Parameters used for evaluation}

For evaluating the performance of YOLOv3 and YOLOv4 used in the RoI extraction stage, various parameters such as Intersection over Union (IOU), precision, recall, F1-score, and mean average precision (mAP) were calculated. The YOLOv3 and YOLOv4 algorithms would identify each lead in the 12-lead ECG image and draw a bounding box over them. These bounding boxes drawn by the YOLOv3 and YOLOv4 would be compared with the bounding boxes which were manually labelled and drawn over the 12-lead ECG images in the dataset. In object detection-based applications, IOU is an important evaluation parameter. The closer the bounding boxes drawn by YOLOv3 and YOLOv4 are to the manually drawn bounding boxes, the higher the value of IOU. The formula for IOU is mentioned in equation \ref{eq1}.

\begin{equation}
    IOU = \frac{Area(Overlap)}{Area(Union)}
    \label{eq1}
\end{equation}

Where,\\
Area(Overlap) refers to the area of the overlap between the bounding boxes drawn by YOLOv3 and YOLOv4 with the manually drawn bounding boxes,\\
Area(Union) refers to the area of the union of the bounding boxes drawn by YOLOv3 and YOLOv4 with the manually drawn bounding boxes.\\

The formulas used for calculating precision, recall, F1-score, and mean average precision (mAP) are mentioned in equation \ref{eq2} to \ref{eq5}, respectively.

\begin{equation}
    Precision = \frac{TP}{TP+FP}
    \label{eq2}
\end{equation}

Where,\\
TP refers to true positives,\\
FP refers to false positives.\\

\begin{equation}
    Recall = \frac{TP}{TP+FN}
    \label{eq3}
\end{equation}

Where,\\
TP refers to true positives,\\
FN refers to false negatives.\\

\begin{equation}
    F1 = 2\times \frac{Precision\times Recall}{Precision + Recall}
    \label{eq4}
\end{equation}

\begin{equation}
    mAP = \frac{1}{n}\sum_{k=1}^{n}AP_{k}
    \label{eq5}
\end{equation}

Where,\\
AP refers to average precision,\\
n refers to number of classes.\\

To evaluate the efficiency of the binarization stage, the binarized output obtained from the pix2pix network was compared to the digitised values manually recorded for each ECG lead image. Accuracy and Root Mean Square Error (RMSE) were used as the evaluation parameters. The output received after binarization using the pix2pix algorithm was in the binary form. So every pixel in the ground truth image and the output generated was compared and true positive, false positive, true negative, and false negative were calculated. Accuracy and RMSE were calculated using the formula mentioned in equation \ref{eq6} and equation \ref{eq7}.

\begin{equation}
    Accuracy = \frac{TP+TN}{TP+FP+TN+FN}
    \label{eq6}
\end{equation}

Where,\\
TP refers to true positives,\\
TN refers to true negatives,\\
FP refers to false positives,\\
FN refers to false negatives.\\

\begin{equation}
    RMSE = \sqrt{\frac{\sum_{i=1}^{N}\left | y(i) - \hat{y}(i) \right |^{2}}{N}}
    \label{eq7}
\end{equation}

Where,\\
y(i) refers to the value of i$^{th}$ pixel in the image generated by the pix2pix model,
$\hat{y}(i)$ refers to the value of i$^{th}$ pixel in the ground truth digitized image,\\
N refers to the total number of pixels.\\

\section{Results} % All tables

\subsection{Region of Intereset}
As previously mentioned, we used 53 images to train the YOLOv3 and YOLOv4 models, out of which 48 were training samples and 5 were testing samples. The 12 lead ECG paper was fed to the YOLOv3 based Darknet53.conv.74 model, which was a pre-trained model. The output of the model provided coordinates of the bounding boxes covering the ECG signal. The coordinates of the corresponding leads were used to crop the lead from the image. The operation was done using Python's OpenCV library. The individual leads numbered 1 to 12 identified by YOLOv3 and YOLOv4 were compared to the manually labelled leads numbered 1 to 12 in the dataset. The performance of YOLOv3 and YOLOv4 was evaluated using mean average precision (mAP) at a 50 \% Intersection Over Union (IOU) threshold. The evaluation was performed on a test dataset consisting of five images, where YOLOv3 had a mAP@50 \% of 0.98 and YOLOv4 had a mAP@50\% of 0.83. A comparison of parameters obtained for YOLOv3 and YOLOv4 is shown in table \ref{tab1}.

\begin{table*}[h!]
%Start from left
%\resizebox{\columnwidth}{!}{%
\centering
\caption{The YOLOv3 and YOLOv4 algorithms are compared. Both the algorithms were tested on the dataset of ECG paper records. The results obtained proved that the YOLOv3 algorithm outperformed the YOLOv4 algorithm with better precision, recall, F1-score, Intersection Over Union (IoU), and Mean Average Precision (mAP).}
\begin{tabular}{lcc}
\hline
\multicolumn{1}{c}{Parameters}       & YOLOv3 & YOLOv4 \\ \hline
Number of training images   & 48     & 48     \\
Number of validation images & 5      & 5      \\
Precision                   & 0.97   & 0.81   \\
Recall                      & 0.94   & 0.81   \\
F1-Score                    & 0.95   & 0.81   \\
Average IoU                 & 0.72   & 0.57   \\
mAP @ 50 \%                    & 0.98   & 0.83   \\ \hline
\end{tabular}
%}

%\label{y3y4comp}
\label{tab1}
\end{table*}

Figure \ref{fig3}  shows the procedure of Region of Interest (ROI) extraction. Figure \ref{fig3}  (a to c) were the mobile captured images of ECG paper records which were fed into the YOLOv3 deep learning based object detection model. The bounding boxes on each lead that were drawn using the coordinates obtained by YOLOv3 are shown in Fig. \ref{fig3}  (d and e). Figure \ref{fig3}  (g to i) shows the single lead images which were cropped using OpenCV.

Figure \ref{fig4}  shows the comparison of samples of ROI extraction using YOLOv3 and YOLOv4. It shows that YOLOv3 had better predictions of bounding boxes compared to YOLOv4.

\subsection{Binarization}
For binarization we have used two models, pix2pix and CycleGAN. We used 76 images for training and 25 images for testing and compared the results of both the models. Figure \ref{fig5}  (a) depicts a single lead ECG picture acquired after detecting the region of interest. Figure \ref{fig5}  (b) depicts an ECG picture that has been preprocessed and binarized with the pix2pix model. Vertical scanning is used in Fig. \ref{fig5}  (c) to recover a 1D signal from the binarized ECG image. Figure \ref{fig5}  (e and f) depicts the extraction of a 1D signal from the blurred ECG picture seen in Fig. \ref{fig5}  (d). Figure \ref{fig5}  (h and i) depicts the extraction of a 1D signal from an ECG image with low contrast and some signal almost overlapping with the background, as shown in Fig. \ref{fig5}  (g). Figure \ref{fig5}  (k) shows the binarized picture created by pix2pix, and Fig. \ref{fig5}  (l) shows the binarized image generated by CycleGAN of the clean ECG image as shown in Fig. \ref{fig5}  (j). Figure \ref{fig5}  (n) depicts the binarized picture produced by pix2pix, whereas Fig. \ref{fig5}  (o) depicts the binarized image produced by CycleGAN of the grayscale ECG image seen in Fig. \ref{fig5}  (m). The results of the pix2pix model were shown in Fig. \ref{fig5}  (a to i). After comparing the results, it was found that the pix2pix algorithm worked better on the ECG images. 

The comparison of outputs of a couple of images generated using pix2pix and CycleGAN was shown in Fig. \ref{fig5}  (j to o). The output generated from the pix2pix model was processed using OpenCV. To get the 1D signal, the output of the binarized ECG image was scanned in the vertical direction.

\section{Discussion and Conclusion}

In the proposed method for ECG digitization, YOLOv3 was used for detecting the region of interest, then the segmented ECG lead was preprocessed and finally binarized using the pix2pix algorithm. Vertical scanning was done to extract the 1D signal. For the region of interest detection, YOLOv4 and YOLOv3 models were trained on the ECG paper records dataset. But YOLOv3 performed better in the detection of ECG signal leads. From the testing dataset, which consisted of 25 images, the YOLOv3 model was able to perform well on 21 images, whereas other images had some background noise and missing signal data. To further test the model, images were converted into grayscale and then tested on the model.

It was found that when the ECG image had a very sharp background, the proposed model failed to predict ECG signals. When the ECG signal on the image was very blurry, certain parts of the signal were not clear. Because of this problem, the proposed model partially failed to binarize such parts. Another approach for binarizing ECG signals was also tried, where CycleGAN was trained on the same training dataset, but it wasn't performing well on the testing dataset. Figure \ref{fig5}  (j to o) shows the comparison of results obtained by pix2pix and CycleGAN, where Fig. \ref{fig5}  (k) and Fig. \ref{fig5}  (n) are the binarized ECG outputs generated by pix2pix, and Fig. \ref{fig5}  (l) and Fig. \ref{fig5}  (o) are the binarized ECG outputs generated by CycleGAN.

The manuscript proposed a method to digitise paper ECG records using deep learning. This method is useful for the digitization of paper medical records, which will help create a digitised register of patient records. The ECG leads are detected first using object detection, after which they are preprocessed and fed into a deep learning model which binarizes the extracted ECG lead. The lead image is then processed to extract a 1D signal using vertical scanning. The deep learning model for binarizing ECG images had an accuracy of 97.4 \%.

\section*{Acknowledgments}
The ECG dataset and diagnostic assistance from Saidhan Hospital and STEMI Global are gratefully acknowledged by the authors.

\section*{Declarations}

\subsection*{Funding information}
No funding was involved in the present work.

\subsection*{Conflicts of interest} 
Authors R. Patil, S. Suraliya, B. Narkhede, S. Varma, R. Patel, V. Sule, and N. Mehendale declare that there has been no conflict of interest.

\subsection*{Code availability}  
The codes will be made available upon reasonable request to the authors.

\subsection*{Authors' contributions}   
Conceptualization was done by R. Patil (RP), S. Suraliya (SS), B. Narkhede (BN), S. Varma (SV) and N. Mehendale (NM). All the literature reading and data gathering were performed by RP, SS, BN, and SV. All the experiments and coding was performed by RP, SS, BN, and SV. The formal analysis was performed by RP, SS, BN, and SV. Manuscript writing original draft preparation was done by SS, BN, SV, V. Sule (VS) and R. S Patel (RSP). Review and editing was done by VS, RSP and NM. Visualization work was carried out by RP, SS, BN, SV, VS, RSP, and NM.

\subsection*{Ethics approval}
Each author consciously ensures that the following conditions are met by the manuscript: 
1) The authors' original work, which has never been published before, is the subject matter of this article. 
2/No other publication of the material is currently being contemplated. 
3) The paper accurately and thoroughly reflects the authors' own research and analysis. 
4). The significant contributions of co-authors and co-researchers are appropriately acknowledged in the work. 
5) The findings are properly contextualised in relation to previous and ongoing research. 
 
\subsection*{Consent to participate}
This article does not contain any studies with animals or humans performed by any of the authors. Informed consent was not required as there were no human participants. All the necessary permissions were obtained from the Institute Ethical Committee and the concerned authorities.

\subsection*{Consent for publication}
Authors have taken all the necessary consents for publication from participants wherever required.

%Bibliography
\bibliographystyle{unsrt}  
\bibliography{references}

\end{document}